\documentclass[preprint]{elsarticle}

\usepackage{lineno,hyperref}
\usepackage{amsmath,amsfonts,amssymb}
\modulolinenumbers[5]

\journal{ }

\usepackage[utf8]{inputenc} 
\usepackage[T1]{fontenc}    
\usepackage{hyperref}       
\usepackage{url}            
\usepackage{booktabs}       
\usepackage{amsfonts}       
\usepackage{nicefrac}       
\usepackage{microtype}      
\usepackage{xcolor}         

\usepackage{array,makecell}
\usepackage{caption}

\usepackage{amsmath,amssymb}

\begin{document}

\begin{frontmatter}

\title{PathFinder: Discovering Decision Pathways in Deep Neural Networks}

\author[1]{Ozan {\.I}rsoy\corref{cor1}} 
\cortext[cor1]{Corresponding author}
\ead{oirsoy@bloomberg.net}
\author[2]{Ethem Alpayd{\i}n}

\address[1]{Bloomberg L.P., 731 Lexington Ave, New York, NY 10022, USA}
\address[2]{Department of Computer Science, {\" O}zye{\u g}in University, {\c C}ekmek{\"o}y 34794, {\.I}stanbul, Turkey}

\begin{abstract}
Explainability is becoming an increasingly important topic for deep neural networks. Though the operation in convolutional layers is easier to understand, processing becomes opaque in fully-connected layers. The basic idea in our work is that each instance, as it flows through the layers, causes a different activation pattern in the hidden layers and in our Paths methodology, we cluster these activation vectors for each hidden layer and then see how the clusters in successive layers connect to one another as activation flows from the input layer to the output. We find that instances of the same class follow a small number of cluster sequences over the layers, which we name ``decision paths." Such paths explain how classification decisions are typically made, and also help us determine outliers that follow unusual paths. We also propose using the Sankey diagram to visualize such pathways. We validate our method with experiments on two feed-forward networks trained on MNIST and CELEB data sets, and one recurrent network trained on PenDigits.
\end{abstract}

\begin{keyword}
Explainability\sep Interpretability\sep Neural Networks\sep Deep Learning\sep Clustering
\end{keyword}

\end{frontmatter}


%
%

\section{Introduction}

Deep neural networks have proven themselves to be impressively accurate in many applications. With wider use of trained models in the real world, interpretability has become an important requirement \cite{Murdoch_2019, yeom_2021, yu_2022}. One of the main drawbacks of neural networks is that they are black-box, that is, the operation carried out in the network, namely, successive layers of nonlinear transformations, is not easy for a human to understand and follow. In the case of a convolutional layer, e.g., in image recognition, the receptive field of a hidden unit covers a predefined patch of the image and its operation is one of template-matching with a filter, which is a simple operator that is relatively easy to understand \cite{visual_zhang}. In the case of fully-connected layers however, where there is no spatial organization and no constraint on connectivity, understanding what is going on and what each hidden unit is responding to becomes difficult. 

In this work, we propose a method named Paths to discover what we call ``decision paths'' that explain the flow of activation as we go from the input layer to the output; we also propose a way to visualize such paths using the Sankey diagram. We explain the method in Section~\ref{sec:method}, discuss our results on two feed-forward networks and one recurrent network in Section~\ref{sec:experiments}, and conclude in Section~\ref{sec:conclusions}.

\section{The Paths Method}
\label{sec:method}

First off, we want to state that our method aims explaining an already trained neural network. We do not modify the network structure, the loss to be optimized, nor the way it is optimized. Whatever the application is, once the network is fully trained after all the usual fine-tuning is done on its hyper-parameters, we pass the training instances in the forward direction through the network one by one and record the activations in all layers. Let $\{x_i\}_i$ denote the set of $i=1,\ldots,N$ instances in our data and for any instance $x_i$, $\{h_i^\ell\}_i$ denote the corresponding hidden unit activation vectors calculated at layer $\ell$, for all hidden layers $\ell$ separately. 

Our work is inspired by the Deep $k$-Nearest Neighbor method ($k$-NN) \cite{deepknn}, which uses the idea that the hidden layers of a deep neural network defines different representations in different spaces. Deep $k$-NN looks at the $k$-nearest neighbors of an instance in these different representation spaces, and whether those neighbors have the same label or not is taken as an indicator of whether that instance is typical or possibly adversarial; such neighbors also allow interpretability, in other words, case-based reasoning.

What we propose is to use clustering on these representations. That is, for each hidden layer $\ell$ separately, we perform clustering on $\{h_i^\ell\}_i^N$. We do this regardless of whether the layer is convolutional or fully-connected, and also regardless of the activation function, ReLU, sigmoid, etc. used at each layer. 

Once clustering is done and the centroids are found, $\{h_i^\ell\}$ are clustered by assigning them to the cluster with the nearest centroid so that $C_j^\ell$ denotes all the instances whose activations in layer $\ell$ fall in cluster $j, j=1,\ldots,k^\ell$, where $k^\ell$ is the number of clusters in layer $\ell$. This 1-of-$k^\ell$ cluster membership information $\{C_j^\ell\}_j$ is much simpler to understand than the typically high-dimensional distributed activation vectors $\{h_i^\ell\}_i$. This is done separately for all layers. 

The number of clusters in the different layers, $k^\ell, \forall l$, are the hyper-parameters that need to be set appropriately. If $k^\ell$ is chosen small, activations that are different would be forced to lie in the same cluster and this would make interpretation difficult; on the other hand, if $k^\ell$ is chosen large, clusters would be divided into many unnecessary sub-clusters and that would make the interpretation unnecessarily complex; but we believe it is better to err on this side of caution and use, or start with, large $k^\ell$.

Once this clustering is done separately for all hidden layers, we can also look at the relationship between clusters in successive layers to tie them together. If many instances that fall in $C_a^\ell$ then later fall in $C_b^{\ell+1}$, this is indicative of a particular flow of decision from cluster $a$ in layer $\ell$ to cluster $b$ in layer $\ell+1$. By looking at such flows from the first layer to the last, we discover ``decision paths.'' Again because of the one-hot nature of clustering, such a relationship between cluster indices in successive layers is easy to interpret whereas the relationship between high-dimensional distributed representations in successive layers is not. 

The clustering operation performs a discretization in the representation space, and given the cluster indices in the different layers we can interpret the network as a machine that moves from one discrete state (a particular cluster index in one layer) to another discrete state (a particular cluster index in the next layer). We can then find such transitions, investigate them, and draw qualitative conclusions.

In trying to understand the transitions between clusters, we make use of the Sankey diagram, traditionally used to visualize flows in a system~\cite{sankey}. It is a network where arrows represent flows and the width of an arrow is proportional to the strength of the flow. In our case, the nodes of the Sankey diagram are clusters organized as layers and the flow between two clusters in two successive layers correspond to the number of training instances that trigger them both. $F_{a,b}^{\ell,\ell+1}$ is the number of training instances that fall in cluster $a$ in layer $\ell$ {\em and\/} fall in cluster $b$ in layer $\ell+1$. The Sankey diagram allows us to visualize these $F_{a,b}^{\ell,\ell+1}$ counts. The main decision pathways can then be read from the diagram by following the wide arrows, and the thin arrows represent the rare instances or outliers.

\section{Experiments}
\label{sec:experiments}

\subsection{Results on MNIST}

\begin{figure}
    \centering
    \includegraphics[width=.7\columnwidth]{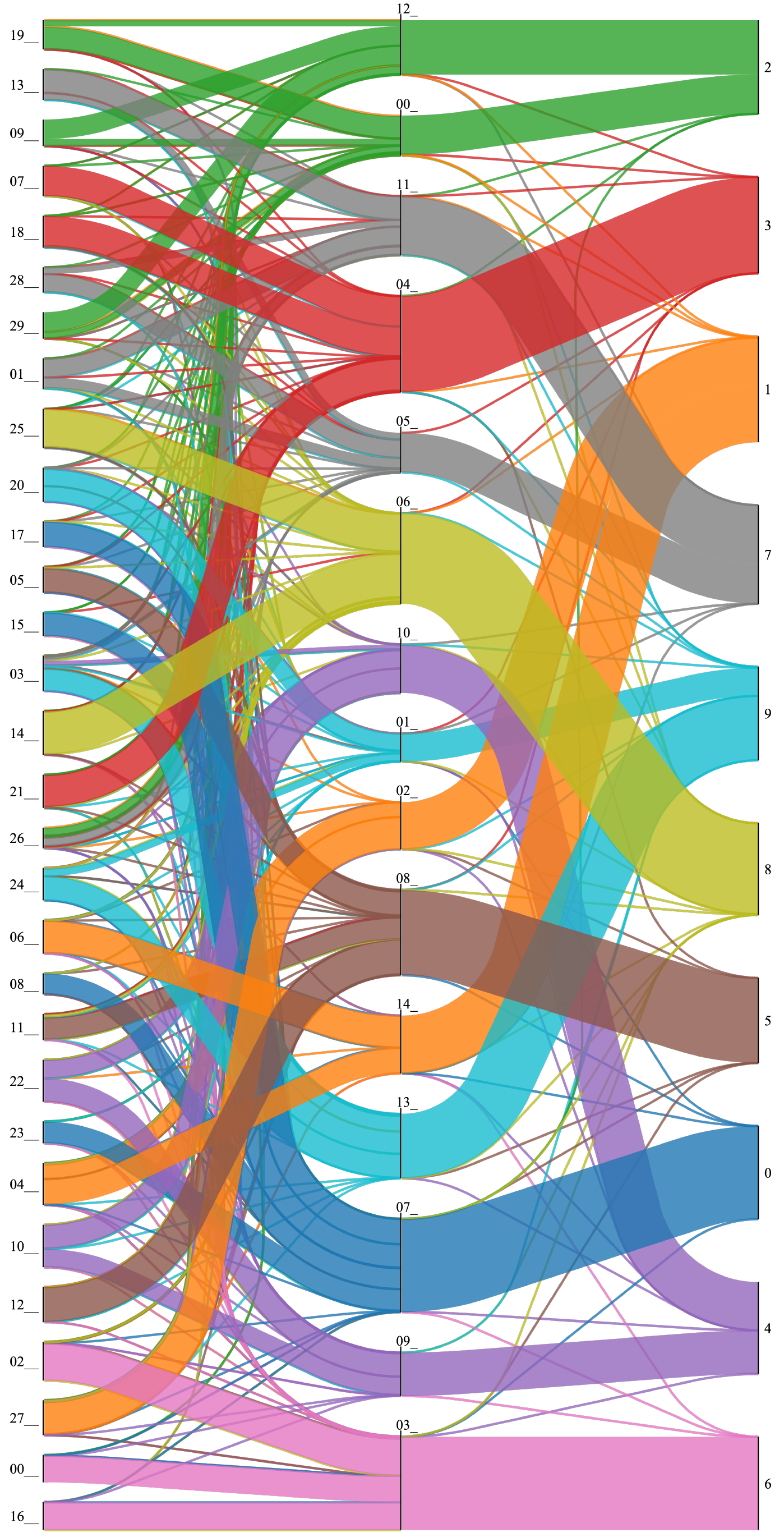}
    \caption{The Sankey diagram for the MNIST network. From left to right, the nodes are the clusters in the second and third hidden layers and then there are the ten nodes for the ten classes. Flows are color-coded using the class value. It can be seen that a large majority of instances for each class are following one of few paths, with thinner flows for atypical cases.}
    \label{fig:mnist-sankey}
\end{figure}

The MNIST data set consists of 70,000 28$\times$28-pixel handwritten digit images belonging to ten classes, which are split into training and test sets as 60,000 and 10,000. We adopt a simple feed-forward architecture with three hidden layers. The first convolutional layer has 32 channels with $5\times 5$ kernels followed by $3\times 3$ max pooling and a stride of 2. The second layer is also convolutional, again with 32 channels with $3\times 3$ kernels, also followed by $3\times 3$ max pooling and a stride of 2. This leads to a 1,568 dimensional representation which feeds to a 100-dimensional tanh hidden layer with fully-connected weights, which then in a fully-connected manner feeds to the ten outputs. We have softmax outputs and minimize cross-entropy using Adam update rule with a learning rate $10^{-3}$. The updates are performed after each mini-batch of size 64. The network for MNIST is trained for 25 epochs. On MNIST, the model is able to achieve 99.25\% accuracy on the test set.

To cluster the representations of each layer, we use $k$-means clustering with Euclidean distance as the metric. The clustering is restarted ten times starting from different random seeds and the solution with the smallest reconstruction error is picked as the final one. 

On MNIST, we use 30 clusters for the second (convolutional) layer and 15 for the third (fully-connected) layer. We do not do any clustering on the activations of the first convolutional layer because we expect basic image filters to be learned there, which we not expect to be informative for classification decisions.

The Sankey diagram is shown in Figure~\ref{fig:mnist-sankey}; we used the \texttt{ipysankeywidget} package to draw the Sankey diagrams \cite{ipysankeywidget}. The nodes on the three columns from left to right correspond to the 30 clusters calculated over the activations of the second (convolutional) layer, 15 clusters calculated over the activations of the third (fully-connected) layer, and the 10 classes. A flow from one node to another corresponds to an instance that activates one cluster in the first and then another in the next, and the width of the flow is proportional to the number of such instances. The flows are colored using ten colors each corresponding to one class. Note that the nodes in the diagram are not sorted by index but are placed to maximize readability. 

We see that most classes are explainable by two or three main paths. For example, following the dark blue flow, we see that instances of class 0 are mostly fed by cluster 7 in the previous layer, which in turn is fed by clusters 17, 15, 8, and 23 in the layer before. Each of these alternatives define one path, which is one different way of being 0. In Table \ref{tab:mnist-paths-0}, we show the average image of instances falling in those clusters. We see that the four paths correspond to four different writing styles for class 0. The last column of Table \ref{tab:mnist-paths-0} show the percentage of each path; for example, the path 17-7-0 cover 27 per cent of the training instances belonging to class 0. In total, these four paths explain more than 97 per cent of all 0 instances. Paths can similarly be extracted for other classes by following their decision paths. 

\begin{table}
\caption{Different paths followed for the class 0 in decreasing order of coverage. The four different writing styles for 0 (shown in dark blue in Figure \ref{fig:mnist-sankey}) are captured by the four clusters, 17, 15, 8, 23, which all feed to cluster 7 in the second layer. These four paths together cover 97.3 per cent of all the instances of class 0.}
\label{tab:mnist-paths-0}
\centering
\begin{tabular}{
    >{\raggedleft\arraybackslash}m{0.5cm}
    >{\centering\arraybackslash}m{2.5cm}
    >{\raggedleft\arraybackslash}m{1cm}
    >{\centering\arraybackslash}m{2.5cm}
    >{\raggedleft\arraybackslash}m{2cm}
    }
    \toprule
	\multicolumn{2}{c}{First cluster} & \multicolumn{2}{c}{Second cluster} \\
    \cmidrule(r){1-2} 
    \cmidrule(r){3-4}
	Index & Average image & Index & Average image & Coverage \\
    \midrule
	    17 & \includegraphics[width=.1\columnwidth]{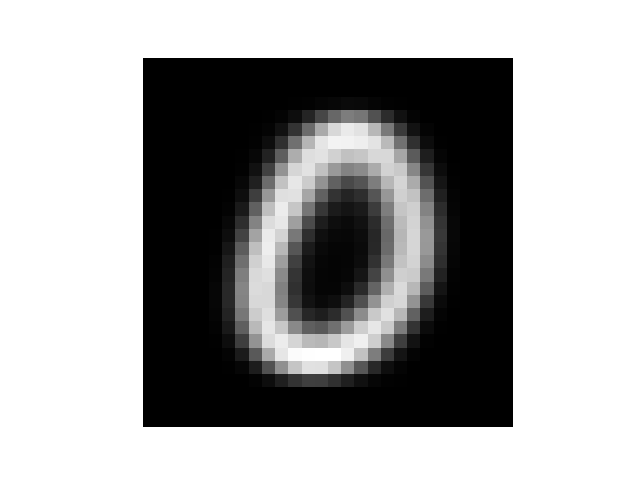} & 
	    7 & \includegraphics[width=.1\columnwidth]{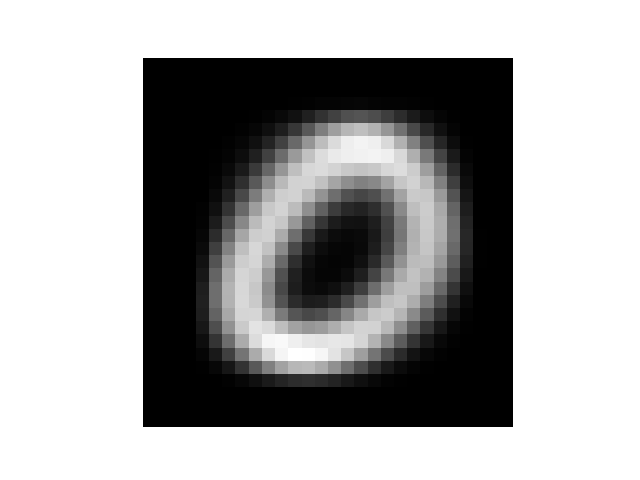} &
    	0.270 \\
        15 & \includegraphics[width=.1\columnwidth]{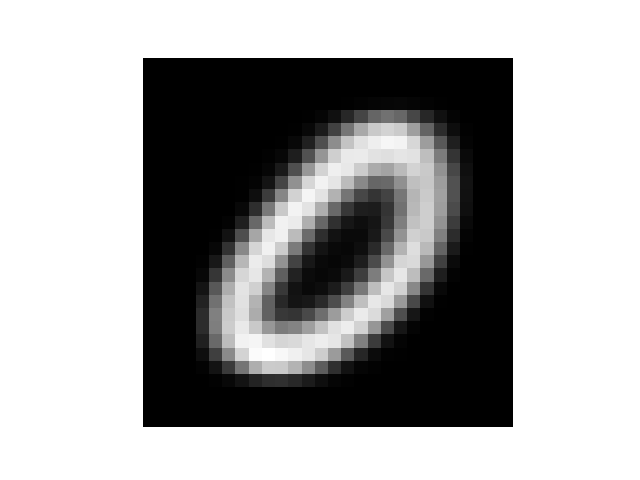} &
   	    7 & \includegraphics[width=.1\columnwidth]{mnist_img/cluster-means/h-07.png} &
    	0.249 \\
        8 & \includegraphics[width=.1\columnwidth]{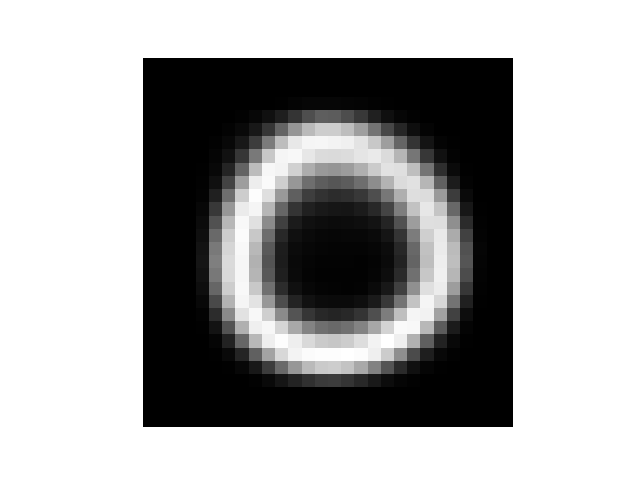} &
   	    7 & \includegraphics[width=.1\columnwidth]{mnist_img/cluster-means/h-07.png} &
    	0.222 \\
        23 & \includegraphics[width=.1\columnwidth]{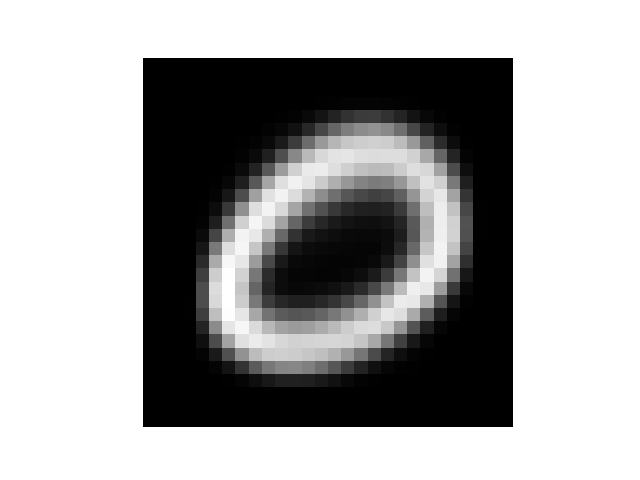} &
   	    7 & \includegraphics[width=.1\columnwidth]{mnist_img/cluster-means/h-07.png} &
    	0.232 \\
    \bottomrule
\end{tabular}
\end{table}

The Sankey diagram allows us to see not only the main pathways (thick flows) but also the rare or exceptional instances (very thin lines). These correspond to instances that follow an unusual path activating different clusters along the way because they are quite different in terms of their representations; examples are shown in Table~\ref{tab:mnist-rare}. For example, at the top row we have instances from cluster 3 in the second level that were assigned to class 0 though that cluster mostly contains instances that were assigned to class 6. In Figure~\ref{fig:mnist-sankey}, cluster 3 is at the bottom of the second column of clusters, and we see that it feeds almost exclusively to class 6; if we look closely, we can also see the thin blue line from cluster 3 to class 0. 

\begin{table}
\caption{Examples of rare transitions.}
\label{tab:mnist-rare}
\centering
\begin{tabular}{
    >{\raggedright\arraybackslash}m{5cm}
    >{\raggedleft\arraybackslash}m{1.5cm}
    >{\raggedleft\arraybackslash}m{1.4cm}
    >{\raggedleft\arraybackslash}m{1cm}
    >{\raggedleft\arraybackslash}m{1.5cm}
    }
    \toprule
    Samples & Second cluster id & Dominant digit & Predicted digit & Coverage \\
    \midrule	
    \includegraphics[height=3em]{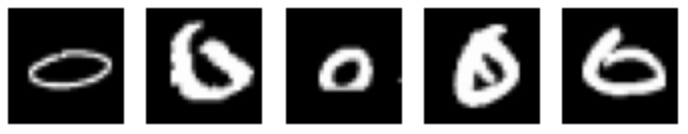} & 3  & 6 & 0 & 0.0014 \\
    \includegraphics[height=3em]{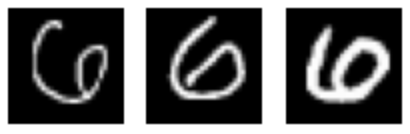} & 7  & 0 & 6 & 0.0005 \\
    \includegraphics[height=3em]{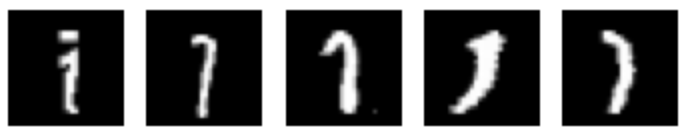} & 11 & 7 & 1 & 0.0007 \\
    \includegraphics[height=3em]{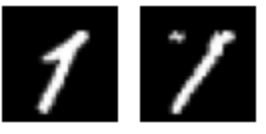} & 14 & 1 & 7 & 0.0003 \\
    \bottomrule
\end{tabular}
\end{table}

\subsection{Results on CelebA}

CelebA consists of $178\times 218$ RGB images of faces with 40 binary labels, which are split into training, validation, and test sets containing 162,770, 19,867, and 19,962 instances respectively~\cite{celeba}. The network has the same architecture used on MNIST except that here there are three RBG channels and the image size is larger. The first convolutional layer has 32 channels with $5\times 5$ kernels followed by $3\times 3$ max pooling and a stride of 2. The second layer is also convolutional, again with 32 channels with $3\times 3$ kernels, also followed by $3\times 3$ max pooling and a stride of 2. On CelebA, this leads to a 20,608 dimensional representation. This layer feeds to a 100-dimensional tanh hidden layer with fully-connected weights, which then in a fully-connected manner feeds to the outputs, which are independent two-class sigmoids for 40-label multi-label classification. Training is done using standard back-propagation with Adam update rule with a learning rate $10^{-3}$. The updates are performed after each mini-batch of size 64. We stop training by looking at the macro-F measure on the validation set. This model achieves 90.2\% accuracy and 64.1\% Macro F-score on the validation set, and 89.7\% accuracy and 63.3\% Macro F-score on the test set.

\begin{figure}
    \centering
    \begin{tabular}{cc}
    (a)		& (b) \\
    \includegraphics[width=.5\columnwidth]{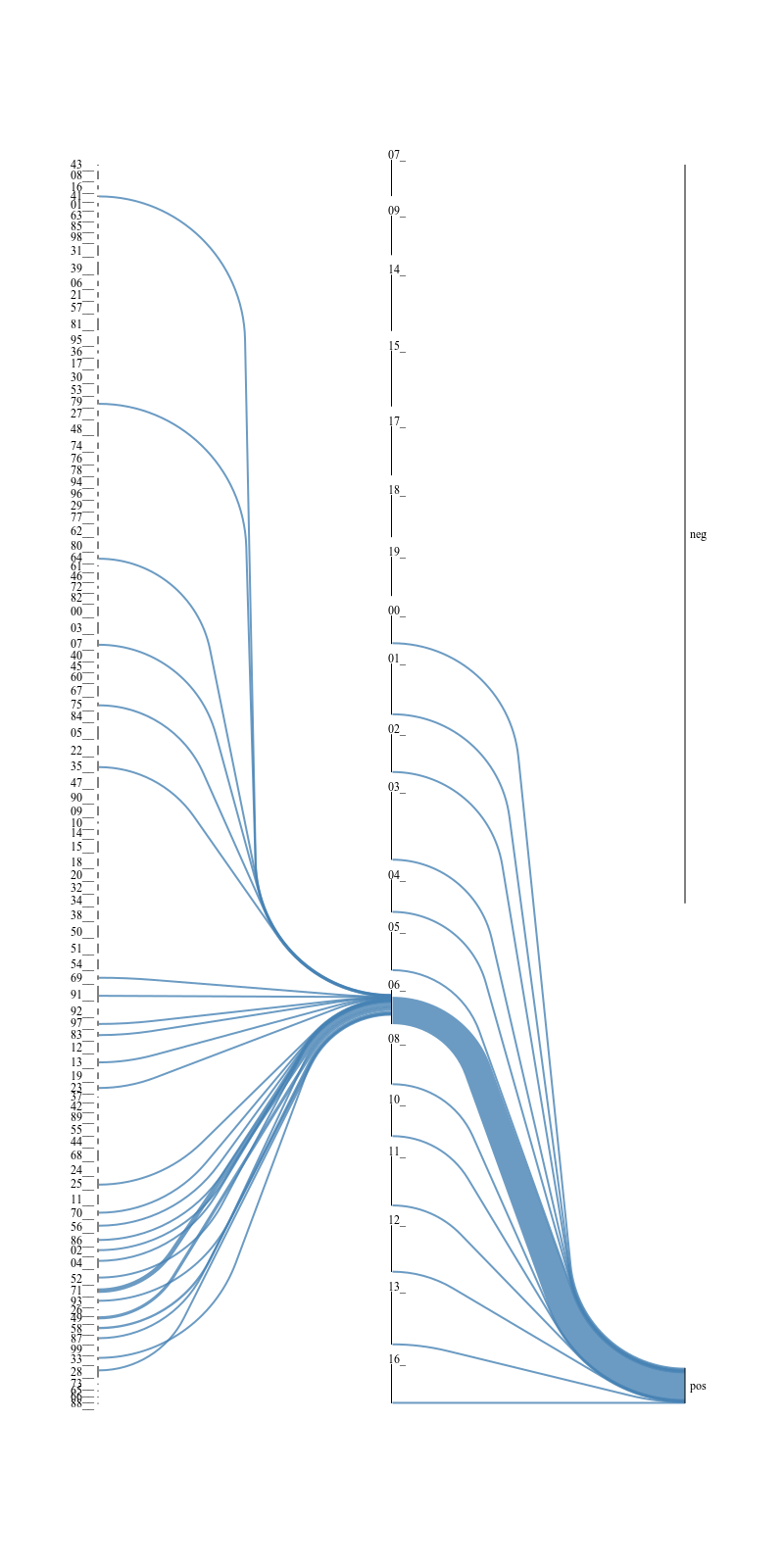} & 
    {\includegraphics[width=.5\columnwidth]{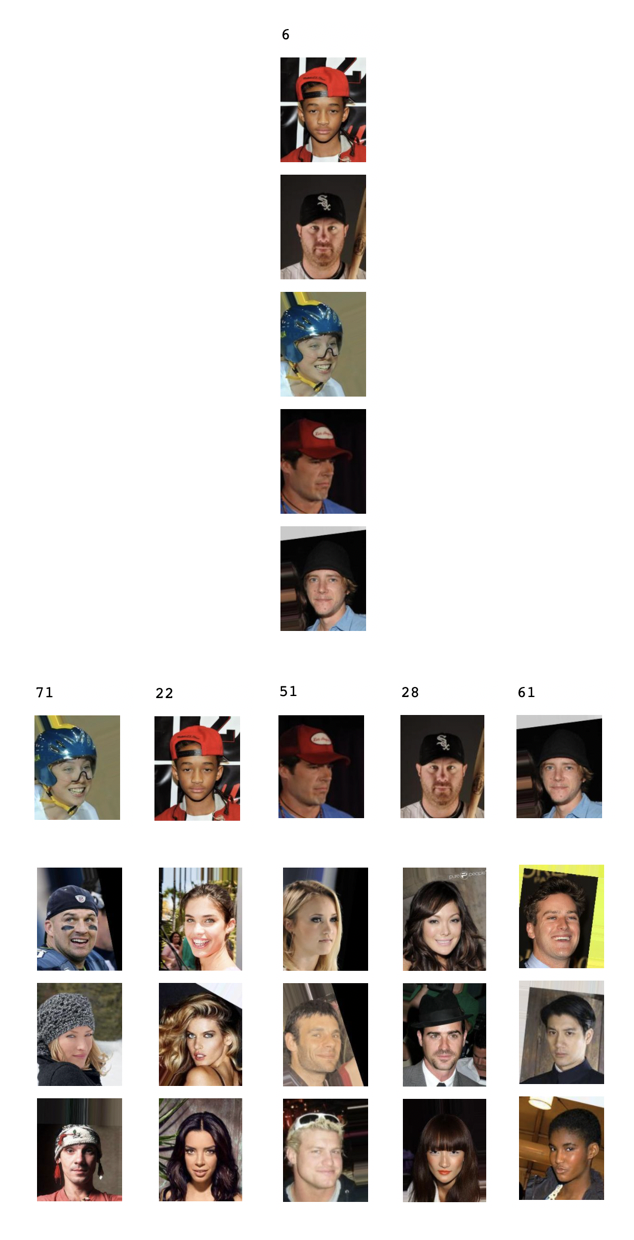}}
    \end{tabular}
    \caption{(a) The Sankey diagram showing the flow for label ``hat'' for a network trained on the CELEB data set, (b) Five random samples falling in cluster 6 in the layer before the output layer, (c) These five samples come from clusters 71, 22, 51, 28 and 61 in the previous layer; three more random samples are shown from each of these clusters to show the type of images falling in each (not necessarily all including hats).}
    \label{fig:celeba-hat}
\end{figure}

On CelebA, we again ignore the first convolutional layer and because of the higher variability than MNIST, we use more clusters, namely, 100 clusters for the second (convolutional) layer and 20 for the third (fully-connected) layer. In Figure \ref{fig:celeba-hat}, we see the Sankey diagram for the ``hat'' label. Cluster 6 seems to be the main cluster in the second cluster layer and in the previous layer, we have clusters for different type of faces feeding to it. Note that though we would expect images falling in cluster 6 to contain images with hats, clusters in the previous layer, closer to the input, would be expected to be more varied in terms of the general image content, and need not all contain images with hats (they may also be part of paths for other labels as well). For example, faces in cluster 51 seem to be looking slightly to the right, containing images with and without hats.

\begin{figure}
    \centering
    \begin{tabular}{cc}
    (a)		& (b) \\
    \includegraphics[width=.5\columnwidth]{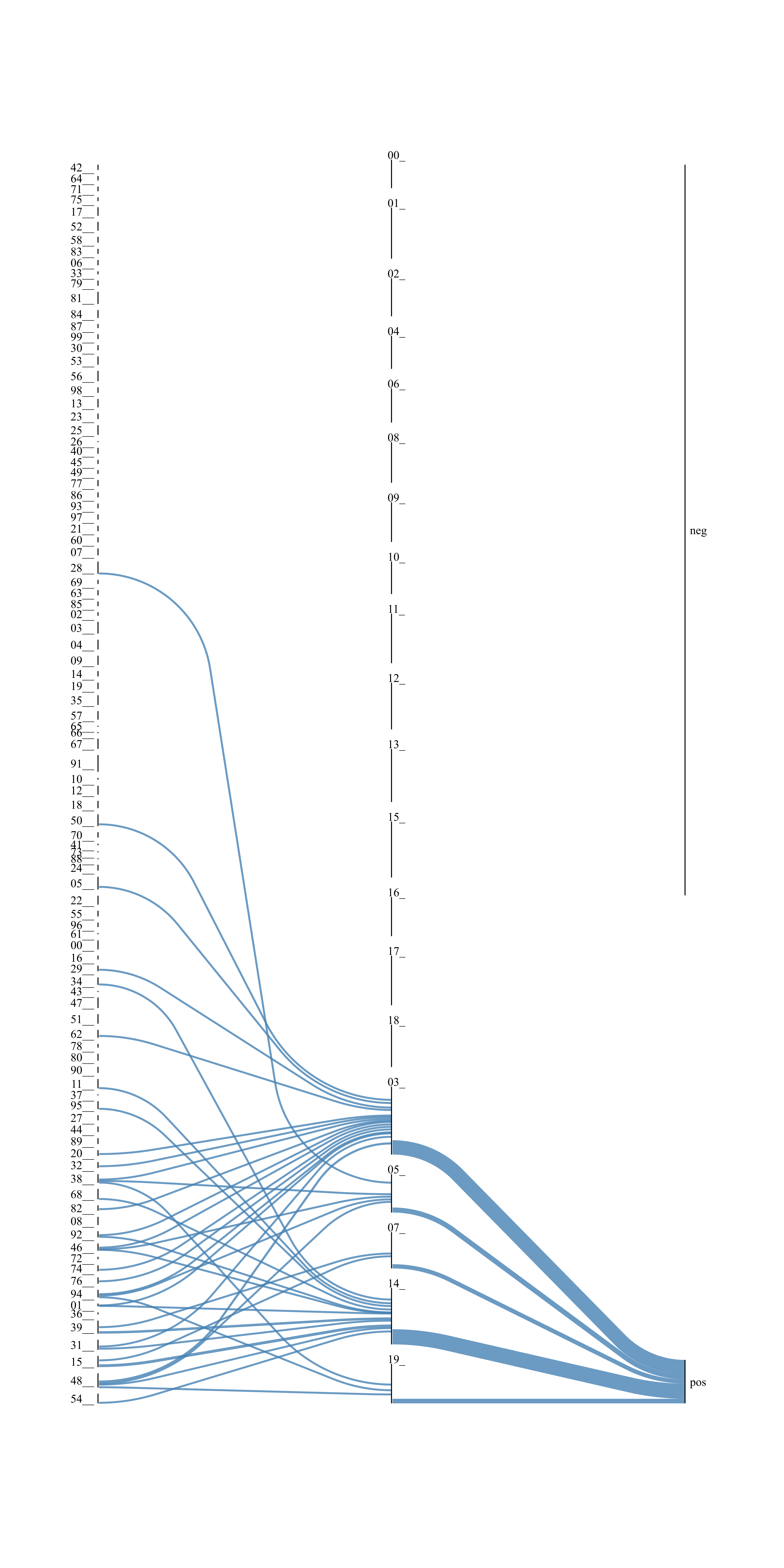} &
    \includegraphics[width=.3\columnwidth]{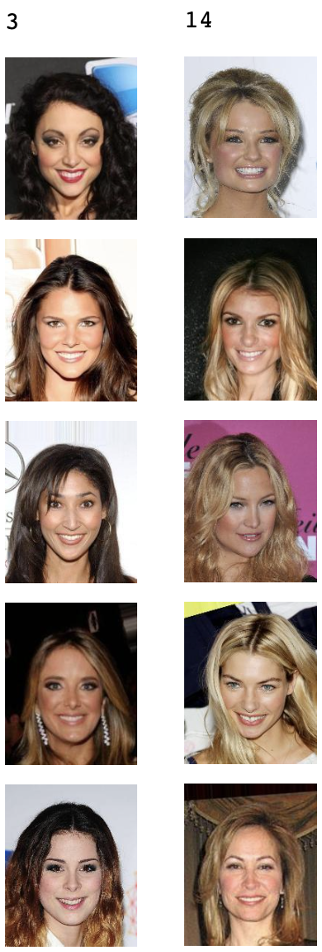}%
    \end{tabular}
    \caption{(a) The Sankey diagram showing the flow for label ``rosy cheeks'' for a network trained on the CELEB data set, (b) Five random samples falling in clusters 3 and 14 in the second cluster layer show that they respond to images with different hair colors.}
    \label{fig:celeba-rosy}
\end{figure}

As another example, in Figure~\ref{fig:celeba-rosy}, we present the diagram for the ``rosy cheeks`` label, which is mainly fed from clusters 3 and 14. When we randomly sample from these clusters with the condition that ``rosy cheeks'' label is assigned, we observe that the two clusters respond to differences due to hair color.

\subsection{Results on PenDigits}

Our method is not applicable just to feedforward networks but can be used for recurrent networks as well. To show that, we have also done experiments using the Pendigits data set, available at the UCI repository, which consists of 5,621 training sequences of 2D coordinates of a stylus on a touch-sensitive tablet while handwriting the ten digits. The writer-dependent validation set contains 1,873 instances (these are samples from writers who have also contributed to the training set), and the writer-independent test set contains 3,498 instances (these are samples from writers who are distinct from writers that contributed to the training set).

For this sequence prediction task, we use the Gated Recurrent Unit (GRU) architecture \cite{gru}. Let $\{x_{i,t}\}_t^{T_i}$ denote a single instance $i$ as a sequence of length $T_i$ and $\{\{x_{i,t}\}_t^{T_i}\}_i^N$ the entire data set of $N$ sequences. Similarly, let $\{\{h_{i,t}\}_t^{T_i}\}_i^N$ denote the set of vector of activations that GRU calculates for each element of every sequence, namely:
\begin{align*}
    h_{i,t} = \text{GRU}(x_{i,t}, h_{i,t-1})
\end{align*}

We use a classification objective where the GRU has an affine-softmax output layer (at the very last time-step) with ten output dimensions and the loss is the cross-entropy. We have a single GRU layer with 50 units. This model achieves 99.7\% accuracy on the writer-dependent validation set  and 92.9\% accuracy on the test set writer-independent accuracy.

As in the previous feed-forward networks, we perform clustering on the representations learned at the hidden layer, that is, $\{\{h_{i,t}\}_t^{T_i}\}_i^N$, which is equivalent to a clustering over all prefix sequences. In this case of sequence prediction, instead of transformations applied from one layer to the next layer, we are focusing on transformations applied from one time-step to the next: Again, we expect that inspecting the transformation between cluster indices, namely, from $\{C_{j,t-1}\}_{j,t}$ to $\{C_{j,t}\}_{j,t}$, would be more interpretable than the transformation between distributed vectors of representations, namely, from $\{h_{i,t-1}\}_{i,t}$ to $\{h_{i,t}\}_{i,t}$. 

\begin{figure}
    \centering
    \includegraphics[width=0.5\textwidth]{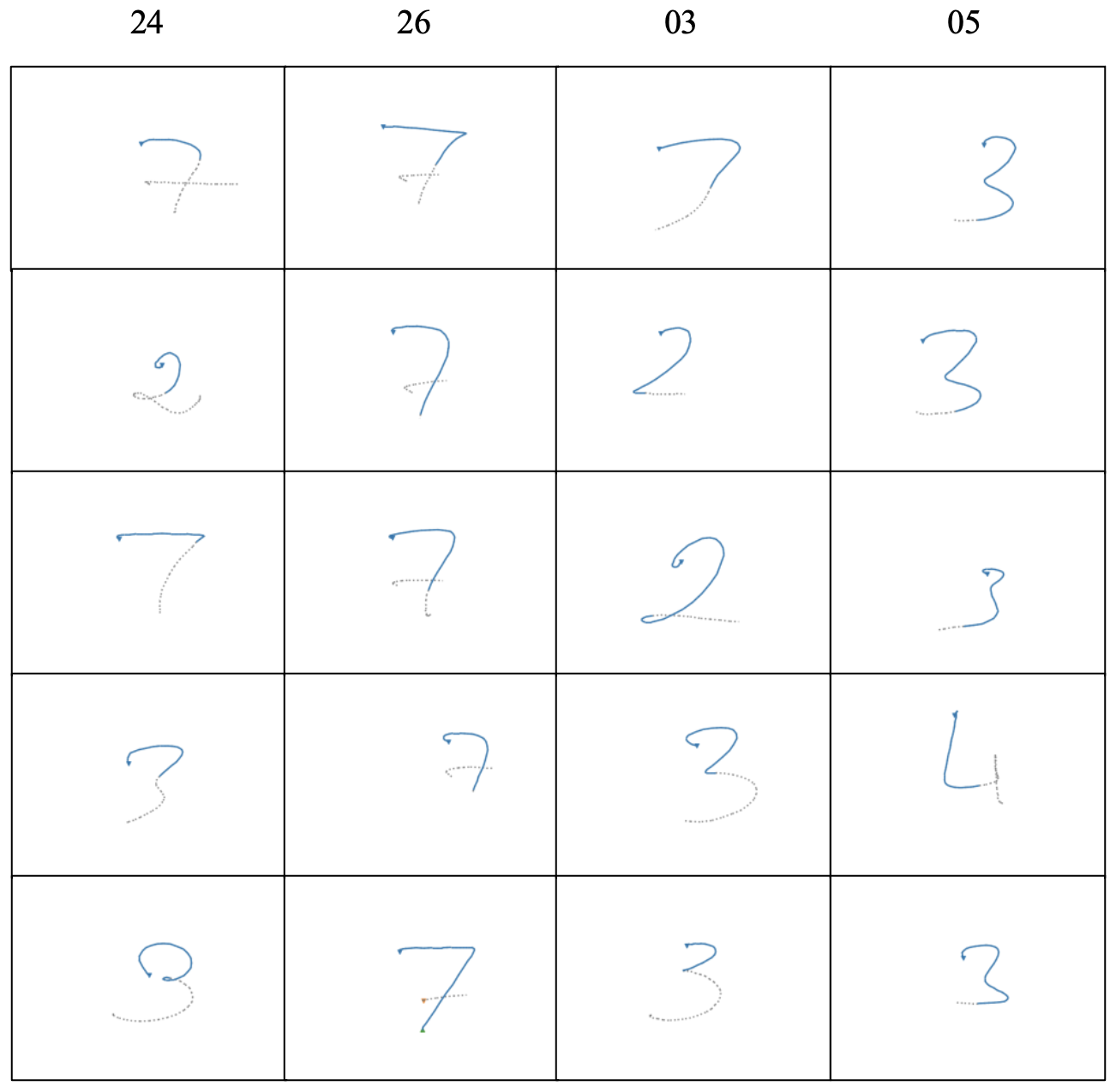}
    \caption{Random samples from clusters 24, 26, 03, and 05, which are clusters traversed by digits 2, 3, and 7 as they are written as a sequence of pen-tip coordinates. The thicker portion of each digit is the part that has been seen until now, the dotted part is the rest yet unseen.}
    \label{fig:clusters-237}
\end{figure}

As an example, a portion of the Sankey diagram is shown in Figure~\ref{fig:pend-237}; the full diagram is in the Supplement. Here, we focus on how digits 2 (green), 3 (red), and 7 (gray) are handled across time-steps 21 through 29. Accompanying Figure~\ref{fig:clusters-237} shows random samples from the four main clusters that are visited by these digits. We observe that early on all three digits utilize cluster 24. As we move in time, instances from class 7 switch to cluster 26. Cluster 24 carries samples that have the top right round stroke that is shared by all three digits. If the stroke progresses downward enough (to distinguish class 7 from 2 or 3), there is a split to cluster 26. Similarly, in time we see instances of classes 2 and 3 shifting from cluster 24 to 03. As time-steps progress even further, instances from class 3 leave cluster 03 for cluster 05. Investigating the random samples from those clusters we observe a similar phenomenon: Cluster 03 examples have yet incomplete strokes that can still be completed into either a 2 or a 3, whereas in cluster 05, we have seen enough to say that they are not of class 2.

\begin{figure}
    \centering
    \includegraphics[width=\textwidth]{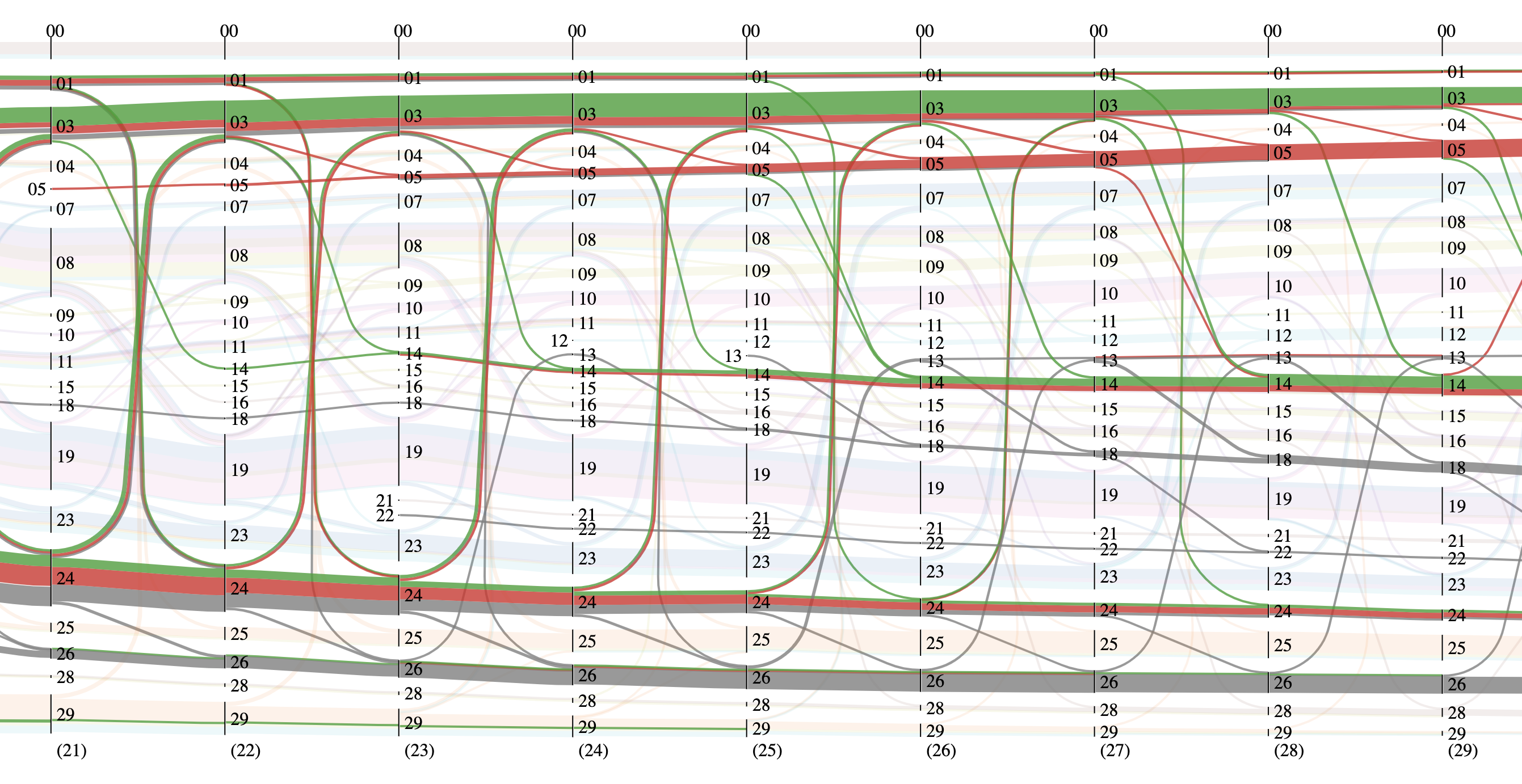}
    \caption{Sankey diagram of cluster transitions across time-steps 21--29, concentrating on classes 2 (green), 3 (red), and 7 (gray). Other digits have their color dimmed to avoid clutter.}
    \label{fig:pend-237}
\end{figure}

\section{Conclusions}
\label{sec:conclusions}

We propose the Paths methodology to extract ``decision paths" from a trained neural network. Such paths explain how classification decisions are typically made, and also help us determine outliers that follow unusual paths. We also propose using the Sankey diagram to visualize such pathways. We validate our method with experiments on two feed-forward networks trained on MNIST and CELEB data sets and one recurrent network trained on PenDigits.

\bibliography{paths_pr.bib}

\end{document}